\pdfoutput=1

\documentclass[11pt]{article}

\usepackage{acl}

\usepackage{times}
\usepackage{latexsym}

\usepackage[T1]{fontenc}

\usepackage[utf8]{inputenc}

\usepackage{microtype}

\usepackage{todonotes}
\usepackage{booktabs, multirow} 
\usepackage{soul}
\usepackage[table]{colortbl} 
\usepackage{changepage,threeparttable} 

\title{Patching Leaks in the Charformer for Efficient Character-Level Generation}

\author{Lukas Edman \qquad  Antonio Toral \qquad Gertjan van Noord \vspace{.2cm}
 \\ Center for Language and Cognition \\ 
 University of Groningen \vspace{.1cm}
 \\ {\tt \small\{j.l.edman, a.toral.ruiz, g.j.m.van.noord\}@rug.nl}
}

\begin{document}
\maketitle
\begin{abstract}
Character-based representations have important advantages over subword-based ones for morphologically rich languages. They come with increased robustness to noisy input and do not need a separate tokenization step.
However, they also have a crucial disadvantage: they notably increase the length of text sequences.
The GBST method from Charformer groups (aka downsamples) characters to solve this, but allows information to leak when applied to a Transformer decoder. 
We solve this information leak issue, thereby enabling character grouping in the decoder.
We show that Charformer downsampling has no apparent benefits in NMT over previous downsampling methods in terms of translation quality, however it can be trained roughly 30\% faster. 
Promising performance on English--Turkish translation indicate the potential of character-level models for morphologically-rich languages.
\end{abstract}

\section{Introduction}




Most state-of-the-art neural machine translation (NMT) systems operate on the subword level,
typically using a preprocessing technique like Byte-Pair Encoding (BPE) \cite{sennrich-etal-2016-neural} for combining characters into subwords. However, using a subword representation often masks important information given by characters, from the syntactic and morphological relatedness of words (e.g., “bake” and “bakes” may be assigned each their own unique token) to devices such as rhyme and alliteration.
Although a byte-level pretrained language model, ByT5 \cite{xue2021byt5}, was more robust to misspellings than its subword counterpart, T5 \cite{raffel2019exploring}, results on common benchmarks such as GLUE did not reflect this inherent advantage. Similarly, in NMT, we have not yet seen advantages from character-level models reflected by their BLEU, chrF, or COMET scores\cite{libovicky2021don}.


The major inhibitor for character-level models is inefficiency, which is due to the longer input and output sequences. For example, the average number of characters per subword for English is around 4 \cite{xue2021byt5}, so the input sequence to a character-level model is effectively 4 times longer. With a Transformer model, the problem is compounded by the complexity of self-attention. To address this, models such as the Charformer \cite{tay2021charformer} introduce a downsampling method prior to the Transformer, which combines characters into pseudo-words, reducing the length of the sequence. The downsampling method used in the Charformer, GBST, was originally intended only for use in the encoder, and recent works attempting to apply the GBST layer to the decoder have failed \cite{libovicky2021don}.\footnote{In consultation with the authors, we noted that their results using GBST in decoding in fact used a different method.} 
Our contributions include:
\begin{enumerate}
    \item We show that there is an information leak in the GBST layer which breaks the typical NMT training scheme of a Transformer model.
    \item We resolve the information leak issue, allowing it to be used in a Transformer decoder.
    \item We provide a simple test to check for information leak in tasks which exhibit causality.
    \item We show that despite Charformer's current popularity, the GBST layer does not perform as well as earlier methods such as \citet{lee2017fully} for NMT.
    \item We give some evidence that character-level models are better for morphologically rich languages.
\end{enumerate}


\section{Patching Information Leaks} \label{sect:leak}


The Charformer modifies ByT5 with the addition of the gradient-based subword tokenization (GBST) layer. This layer mixes character representations (which are also informed of their relative positions via a convolution) using a simple mean computed over character n-grams up to length 4.\footnote{For decoding, we include n-grams up to the length of the downsampling factor.} The model then selects a weighted average of these representations before a downsampling via a block-wise mean pooling. The result is a sequence of pseudo-word-level embeddings that is fed in directly to the Transformer (in contrast with the standard sequence of subword embeddings). 

The GBST layer can be applied directly to a Transformer encoder without issue, but it cannot be applied to a Transformer decoder for generative tasks. This is due to an information leak, where information about characters in future blocks can end up in prior blocks. This occurs for 2 reasons: the convolution used to inform position, and the character n-gram means, both of which can overlap with the block-wise separations. 

Concerning the convolution, each character is informed by its neighboring characters, which is necessary for the convolution to serve as a positional embedding. However, this creates the issue that characters on the right side of a block will be informed by characters to the left in a future block.  

\begin{figure}
    \centering
    \includegraphics[scale=0.45]{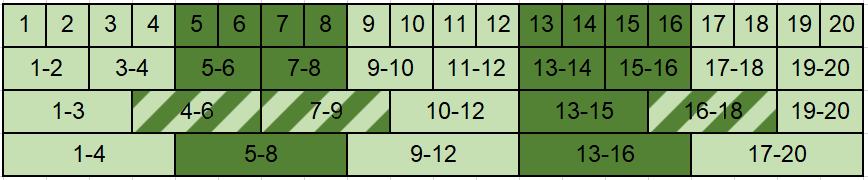}
    \caption{Example of the mean mixing in the GBST layer, with alternating block colors assigned according to what information is accessible via mean mixing. Numbers indicate the range of positions accessible to each n-gram. The striped n-grams indicate a source of information leak.}
    \label{fig:leak}
\end{figure}

The character n-gram averaging similarly can obtain information from future blocks. As seen in Figure \ref{fig:leak}, with a downsampling factor of 4, the issue occurs with trigrams, where the 4th position is averaged with the 5th and 6th, despite being in separate blocks. Therefore, when the Transformer must predict the block containing the 5th and 6th characters, it has already received information about them, and thus can learn to simply copy the characters. 

\subsection{A Simple Test for Information Leaks}

To confirm that there is a leak in the GBST layer, we set up a simple model consisting of a GBST layer, followed by an upsampling layer, which is a linear layer that takes the downsampled block representation and upsamples it back to characters. We then train the model to predict a sequence of random characters, conditioned on the prior characters in the sequence, using a left-padding of BOS tokens equal to the downsampling factor. (e.g., ‘[BOS] [BOS] [BOS] a b c’ is used to predict ‘a b c d e f’).
We expect that, if there is no information leak, the model will have a near-random accuracy, and conversely if the accuracy is significantly higher than random, there is an information leak. 

\begin{table*}[!htp]\centering
\scriptsize
\begin{tabular}{lrrrrrrrrrrrrrr}\toprule
$\delta$ &Pos Embs &1 &2 &3 &4 &5 &6 &7 &8 &9 &10 &11 &12 \\\midrule
\multirow{2}{*}{2} &Sin &\cellcolor[HTML]{ffffff}0.0103 &\cellcolor[HTML]{ffffff}0.0109 &\cellcolor[HTML]{ffffff}0.0103 &\cellcolor[HTML]{ffffff}0.0122 &\cellcolor[HTML]{ffffff}0.0106 &\cellcolor[HTML]{ffffff}0.0113 &\cellcolor[HTML]{ffffff}0.0141 &\cellcolor[HTML]{ffffff}0.0122 &\cellcolor[HTML]{ffffff}0.0147 &\cellcolor[HTML]{ffffff}0.0116 &\cellcolor[HTML]{ffffff}0.0113 &\cellcolor[HTML]{ffffff}0.0075 \\
&Conv &\cellcolor[HTML]{e06666}1.0000 &\cellcolor[HTML]{ffffff}0.0116 &\cellcolor[HTML]{e06666}0.9941 &\cellcolor[HTML]{ffffff}0.0097 &\cellcolor[HTML]{e06666}0.9947 &\cellcolor[HTML]{ffffff}0.0100 &\cellcolor[HTML]{e06666}0.9959 &\cellcolor[HTML]{ffffff}0.0113 &\cellcolor[HTML]{e06666}0.9966 &\cellcolor[HTML]{ffffff}0.0097 &\cellcolor[HTML]{e06666}0.9962 &\cellcolor[HTML]{ffffff}0.0113 \\
\multirow{2}{*}{3} &Sin &\cellcolor[HTML]{e06666}0.1356 &\cellcolor[HTML]{ffffff}0.0078 &\cellcolor[HTML]{ffffff}0.0109 &\cellcolor[HTML]{ffffff}0.0100 &\cellcolor[HTML]{ffffff}0.0078 &\cellcolor[HTML]{ffffff}0.0103 &\cellcolor[HTML]{e06666}0.0241 &\cellcolor[HTML]{ffffff}0.0106 &\cellcolor[HTML]{ffffff}0.0088 &\cellcolor[HTML]{ffffff}0.0091 &\cellcolor[HTML]{ffffff}0.0103 &\cellcolor[HTML]{ffffff}0.0081 \\
&Conv &\cellcolor[HTML]{e06666}1.0000 &\cellcolor[HTML]{e06666}1.0000 &\cellcolor[HTML]{ffffff}0.0072 &\cellcolor[HTML]{e06666}0.9987 &\cellcolor[HTML]{e06666}0.9981 &\cellcolor[HTML]{ffffff}0.0106 &\cellcolor[HTML]{e06666}0.9772 &\cellcolor[HTML]{e06666}0.9619 &\cellcolor[HTML]{ffffff}0.0131 &\cellcolor[HTML]{e06666}0.9966 &\cellcolor[HTML]{e06666}0.9981 &\cellcolor[HTML]{ffffff}0.0081 \\
\multirow{2}{*}{4} &Sin &\cellcolor[HTML]{e06666}0.1347 &\cellcolor[HTML]{e06666}0.0338 &\cellcolor[HTML]{ffffff}0.0137 &\cellcolor[HTML]{ffffff}0.0113 &\cellcolor[HTML]{e06666}0.0262 &\cellcolor[HTML]{ffffff}0.0106 &\cellcolor[HTML]{ffffff}0.0122 &\cellcolor[HTML]{ffffff}0.0100 &\cellcolor[HTML]{ffffff}0.0113 &\cellcolor[HTML]{ffffff}0.0097 &\cellcolor[HTML]{ffffff}0.0063 &\cellcolor[HTML]{ffffff}0.0072 \\
&Conv &\cellcolor[HTML]{e06666}1.0000 &\cellcolor[HTML]{e06666}0.9994 &\cellcolor[HTML]{e06666}1.0000 &\cellcolor[HTML]{ffffff}0.0119 &\cellcolor[HTML]{e06666}0.9775 &\cellcolor[HTML]{e06666}0.9744 &\cellcolor[HTML]{e06666}0.9781 &\cellcolor[HTML]{ffffff}0.0137 &\cellcolor[HTML]{e06666}0.9966 &\cellcolor[HTML]{e06666}0.9900 &\cellcolor[HTML]{e06666}0.9950 &\cellcolor[HTML]{ffffff}0.0122 \\
\bottomrule
\end{tabular}
\caption{Accuracies obtained by the simple model for each position in a target sequence of 12 random tokens, with varying downsampling factor ($\delta$) and either convolutional or sinusoidal positional embeddings. The white boxes denote accuracies within random chance ($p>1e-3$) and thus not leaking, while red denotes accuracies that are significantly higher than random chance ($p<1e-10$), showing an information leak. }\label{tab:leaks}
\end{table*}

In Table \ref{tab:leaks}, we show the specific positions in a sequence affected by the leaks. Details of these experiments can be found in Appendix \ref{A:leak}. We use a vocabulary of size 100, so the accuracy of a perfectly random classifier should be around 1\%, however we see in several cases that the models perform significantly better. In the worst case, with a downsampling factor of 4 and using the convolutional positional embeddings, 75\% of the tokens are leaked. 

To resolve this issue, we considered three potential solutions: adding more padding, removing leaking n-grams, or applying a mask to the n-grams.

\paragraph{Approach 1: Additional Padding.}
The simplest approach is to pad the sequences enough that leaking is impossible. Preliminary experiments indicated that padding with 2 times the downsampling factor prevented any leak. This approach however implies that the predictions are no longer conditioned on the block 1 step back, but rather the block 2 steps back. 

\begin{figure}
    \centering
    \includegraphics[scale=0.45]{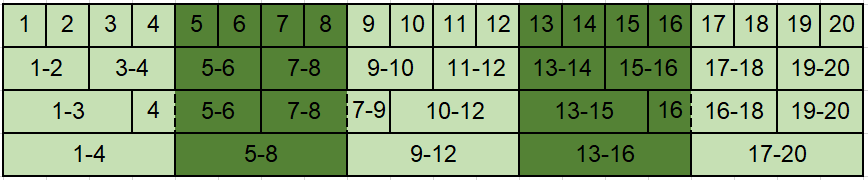}
    \caption{Our third proposed solution to preventing information leak.}
    \label{fig:leak_sol}
\end{figure}

\paragraph{Approach 2: Remove Overlapping N-Grams.}
Our second approach instead addresses the sources of information leak. First, the convolution used as a positional embedding is replaced with a static sinusoidal positional embedding, as in \citet{vaswani2017attention}.  Second, the means for n-grams where there is overlap are removed. For example, for a downsampling factor of 4, the means for trigrams are removed. These changes prevent information leak, but it is possible that the removal of trigrams or other n-grams will substantially worsen the model’s ability to learn contextual character embeddings before downsampling into blocks.

\paragraph{Approach 3: Apply Causal Mask.}
Our third approach builds on our second approach by keeping the sinusoidal positional embeddings and, rather than removing problematic n-grams, applies a causal mask when computing the mean. This can be seen in Figure \ref{fig:leak_sol}, where overlapping n-grams are split by block, with the right side being informed by the left, but the left side being uninformed by the right. This approach allows trigram information to remain when it is not overlapping, however the information present in the blocks is not always consistent. This  might be difficult for the model to discern.

\section{Experimental Setup} \label{sect:exp}
We test our models on English translation to and from Arabic and German, as well as Turkish, whose vowel harmony and agglutination should benefit from character-level processing. Details of the datasets, preprocessing, and hyperparameters can be found in Appendix \ref{A:exp}.  

We compare performance of our modified GBST methods to earlier works from \citet{lee2017fully}. \citet{lee2017fully} found a performant method of character-level MT using a downsampling method based on causal convolutions, followed by a max-pooling layer. Their original MT model used an RNN, and  \citet{libovicky2021don} modernized the model to use a Transformer, keeping the convolution-based downsampling. It notably performed the best in \citet{libovicky2021don}'s experiments on various downsampling methods. We similarly use the Transformer version in our experiments.

For decoding with \citeauthor{lee2017fully}'s and the Charformer downsampling methods, we apply the two-step decoder of \citet{libovicky2021don}, which adds an LSTM layer to the head of the Transformer, receiving its hidden states in addition to character embeddings of the so-far generated output sequence, and outputting the next characters. The parameters used for \citeauthor{lee2017fully}'s method follow \citet{libovicky2021don}.

For evaluation we use BLEU \cite{papineni-etal-2002-bleu} and COMET \cite{rei-etal-2020-comet} with the reference-based \verb|wmt20-comet-da| model. Our code is made freely available.\footnote{\url{https://github.com/Leukas/PatchingGBST}} 

\section{Results} \label{sect:res}
We compare the performance of our modified GBST layers in Table \ref{tab:gbst}. First, we observe the GBST with additional padding, like the non-causal GBST, fails to learn any form of translation. This indicates that predicting two blocks into the future is too difficult a task for the model to make any meaningful progress in learning to translate.

\begin{table}[!htp]\centering
\scriptsize
\begin{tabular}{lrrrrr}\toprule
$\delta$ & Non-Causal &Padding &Removal &Masking \\\midrule
2 & 0.00 & 0.04 &25.49 &24.59 \\
3 & 0.00 & 1.35 &24.90 &22.16 \\
4 & 0.00 &0.62 &22.07 &17.47 \\ \midrule
2 & 0.00 & 0.80 &22.17 &21.90 \\
3 & 0.00 & 0.52 &20.93 & 19.07 \\
4 & 0.00 &0.57 &20.25 &18.04 \\ 

\bottomrule
\end{tabular}
\caption{BLEU scores of vanilla GBST (Non-Causal) and our 3 approaches on German$\rightarrow$English (top) and English$\rightarrow$German (bottom). The same trend was observed for Arabic$\leftrightarrow$English, as seen in Appendix \ref{A:mods}.}\label{tab:gbst}
\end{table}

We also see that the simpler approach of dropping overlapping n-grams works better than applying a causal mask. We suspect this is due to a lack of consistency within the masked n-gram representations, as some are simply duplicates of the lower order n-gram representations. 

The downsampling factor also plays no role in the difference between the methods. Although the average length of a subword in English and German is close to 4 characters, making a downsampling factor of 4 an intuitive choice, a higher downsampling factor leads to worse performance.

\begin{table}[!htp]\centering
\tiny
\begin{tabular}{lrrrrrrrrr}\toprule
Method&$\delta$ &de-en &en-de &ar-en &en-ar &tr-en &en-tr &avg \\\midrule

Subword &1 &27.2 &24.1 &25.6 &\textbf{11.2} &14.7 & 11.3 &\textbf{19.0} \\\midrule
Char &1 &\textbf{27.4} &\textbf{24.3} &\textbf{26.3} &8.7 &\textbf{15.5} &\textbf{11.9} &\textbf{19.0} \\ \midrule
\multirow{2}{*}{r-GBST} &2 &24.8 &22.2 &24.2 &7.8 &14.2 &10.6 &17.3 \\
&4 &22.1 &20.2 &22.2 &8.3 &11.3 &8.2 &15.4 \\ \midrule
\multirow{2}{*}{Lee} &2 &27.1 &23.3 &25.1 &9.9 &15.2 &11.1 &18.6 \\
&4 &24.3 &21.1 &22.6 &9.2 &12.8 &9.3 &16.6 \\

\bottomrule
\end{tabular}
\vspace{1em}

\tiny
\begin{tabular}{lrrrrrrrrr}\toprule
Method&$\delta$ &de-en &en-de &ar-en &en-ar &tr-en &en-tr &avg \\\midrule

Subword &1 &0.20 &\textbf{0.07} &0.06 &\textbf{0.12} &-0.10 &0.18 &\textbf{0.09} \\ \midrule
Char &1 &\textbf{0.22} &0.04 &\textbf{0.14} &-0.01 &\textbf{-0.03} &\textbf{0.19} &\textbf{0.09} \\ \midrule
\multirow{2}{*}{r-GBST} &2 &0.11 &-0.10 &0.03 &-0.11 &-0.13 &0.04 &-0.03 \\ 
&4 &-0.12 &-0.35 &-0.14 &-0.14 &-0.35 &-0.21 &-0.22 \\ \midrule
\multirow{2}{*}{Lee} &2 &\textbf{0.22} &-0.02 &0.07 &0.04 &-0.05 &0.18 &0.07 \\
&4 &0.02 &-0.24 &-0.07 &-0.04 &-0.22 &-0.00 &-0.09 \\

\bottomrule
\end{tabular}
\caption{BLEU (top) and COMET (bottom) scores. ``Char'' refers to a character-level model without any downsampling, with the same architecture and training scheme as ``Subword'', differing only in tokenization. Best results in bold.}\label{tab:main}
\end{table}

Despite the modified GBST 
with n-grams removed (henceforth r-GBST) being the best performing modification, Table \ref{tab:main} shows that it is still outperformed by the downsampling method from \citet{lee2017fully}. The mode of operation between \citeauthor{lee2017fully}'s method and the GBST method is similar: both achieve a mixture of characters, focusing only on neighboring characters to reduce computational complexity. While GBST achieves this with averaging across unigrams to 4-grams, \citeauthor{lee2017fully} uses convolutions with differing kernel sizes. The mixing via convolution is more uniform in nature; for example, the convolution with kernel size 3 is analogous to the trigram mixing, however the convolution does not have a hard boundary after every 3rd character. This may be the reason for \citeauthor{lee2017fully}'s method's superior performance. 

These results raise the question of the significance of the performance of the Charformer for any NLP task. If \citeauthor{lee2017fully}'s method was used in the same pretraining setup used in \citet{tay2021charformer}, would we perhaps see superior performance?

Looking at the results with respect to language pairs, the results on English--Turkish seem most promising for character-level models.
The character-level model with no downsampling outperforms the subword-level in both directions, Lee's method with $\delta=2$ is competitive, and r-GBST with $\delta=2$ is somewhat competitive into-English. This could indicate that models with access to character-level information benefit when translating both to and from languages with rich morphology such as Turkish, although more extensive testing on other morphologically rich languages is needed. 

\begin{table}[!htp]\centering
\scriptsize
\begin{tabular}{lrrrrrr}\toprule
Method &Decoding & $\delta$ & Epoch & Completion &Eval \\\midrule
\multirow{2}{*}{Subword} &Normal &1 &5:34 &3:52:31 &1:33 \\
&2-Step &1 &5:47 &5:15:57 &1:40 \\ \midrule
\multirow{2}{*}{Char} &Normal &1 &14:37 &14:52:21 &5:15 \\
&2-Step &1 &16:45 &17:02:35 &5:19 \\ \midrule
\multirow{2}{*}{r-GBST} &\multirow{2}{*}{2-Step} &2 &10:38 &10:15:02 &3:19 \\
& &4 &8:08 &8:16:32 &2:03 \\ \midrule
\multirow{2}{*}{Lee} &\multirow{2}{*}{2-Step} &2 &13:09 &13:22:48 &5:45 \\ 
& &4 &10:07 &11:55:09 &2:43 \\
\bottomrule
\end{tabular}
\caption{Average time to train for 1 epoch and to completion, and to evaluate on the German$\rightarrow$English dataset (hours + minutes + seconds). Times for other language pairs are proportionally similar.}\label{tab:time}
\vspace{-4mm} 
\end{table}

Both downsampling methods perform worse than using no downsampling, but there is a trade-off in training time. In Table \ref{tab:time}, we show the time it takes to train and evaluate on the test set post-training. We include character and subword-level models where we use the two-step decoding method, in order to separate the effect of downsampling from the decoding method.\footnote{The performance of the 2-step character and subword models are similar to their normal counterparts.} We can see that our downsampling methods are faster for both training and generation than the character model. However the subword model is still the fastest and achieves the best performance.

\section{Conclusion}
\label{sect:concl}
\vspace{-1mm} 

Character-level or byte-level models are intuitive for a variety of reasons but are slower due to much longer sequences. Downsampling methods such as the GBST layer in the Charformer looked promising, but is not usable for generative tasks without modification due to a leak of information flowing from future blocks. With modification, the GBST layer does not perform as well as older methods such as that of \citet{lee2017fully}, although it is faster in both training and evaluation.

Despite the intuitiveness of downsampling from characters to pseudo-words, we see a clear trade-off of performance versus time, with performance decreasing by several BLEU points, but the training and generation time being reduced to up to 50\% of the non-downsampled model. 

Although both downsampling methods tested do not reach the performance of the standard character-level model, subword-level models show that shortening the sequence length can lead to appreciably faster models without any sacrifice in performance. Thus some form of downsampling is beneficial, and our method to find  information leaks can serve as a useful debugging tool.

Finally, the results for English--Turkish show a positive effect of character-level information for translation of morphologically rich languages. 

\bibliography{anthology,custom}

\begin{thebibliography}{12}
\expandafter\ifx\csname natexlab\endcsname\relax\def\natexlab#1{#1}\fi

\bibitem[{Kingma and Ba(2014)}]{kingma2014adam}
Diederik~P Kingma and Jimmy Ba. 2014.
\newblock Adam: A method for stochastic optimization.
\newblock \emph{arXiv preprint arXiv:1412.6980}.

\bibitem[{Kudo and Richardson(2018)}]{kudo2018sentencepiece}
Taku Kudo and John Richardson. 2018.
\newblock Sentencepiece: A simple and language independent subword tokenizer
  and detokenizer for neural text processing.
\newblock \emph{arXiv preprint arXiv:1808.06226}.

\bibitem[{Lee et~al.(2017)Lee, Cho, and Hofmann}]{lee2017fully}
Jason Lee, Kyunghyun Cho, and Thomas Hofmann. 2017.
\newblock Fully character-level neural machine translation without explicit
  segmentation.
\newblock \emph{Transactions of the Association for Computational Linguistics},
  5:365--378.

\bibitem[{Libovick{\`y} et~al.(2021)Libovick{\`y}, Schmid, and
  Fraser}]{libovicky2021don}
Jind{\v{r}}ich Libovick{\`y}, Helmut Schmid, and Alexander Fraser. 2021.
\newblock Why don't people use character-level machine translation?
\newblock \emph{arXiv preprint arXiv:2110.08191}.

\bibitem[{Loshchilov and Hutter(2017)}]{loshchilov2017decoupled}
Ilya Loshchilov and Frank Hutter. 2017.
\newblock Decoupled weight decay regularization.
\newblock \emph{arXiv preprint arXiv:1711.05101}.

\bibitem[{Papineni et~al.(2002)Papineni, Roukos, Ward, and
  Zhu}]{papineni-etal-2002-bleu}
Kishore Papineni, Salim Roukos, Todd Ward, and Wei-Jing Zhu. 2002.
\newblock \href {https://doi.org/10.3115/1073083.1073135} {{B}leu: a method for
  automatic evaluation of machine translation}.
\newblock In \emph{Proceedings of the 40th Annual Meeting of the Association
  for Computational Linguistics}, pages 311--318, Philadelphia, Pennsylvania,
  USA. Association for Computational Linguistics.

\bibitem[{Raffel et~al.(2019)Raffel, Shazeer, Roberts, Lee, Narang, Matena,
  Zhou, Li, and Liu}]{raffel2019exploring}
Colin Raffel, Noam Shazeer, Adam Roberts, Katherine Lee, Sharan Narang, Michael
  Matena, Yanqi Zhou, Wei Li, and Peter~J Liu. 2019.
\newblock Exploring the limits of transfer learning with a unified text-to-text
  transformer.
\newblock \emph{arXiv preprint arXiv:1910.10683}.

\bibitem[{Rei et~al.(2020)Rei, Stewart, Farinha, and
  Lavie}]{rei-etal-2020-comet}
Ricardo Rei, Craig Stewart, Ana~C Farinha, and Alon Lavie. 2020.
\newblock \href {https://doi.org/10.18653/v1/2020.emnlp-main.213} {{COMET}: A
  neural framework for {MT} evaluation}.
\newblock In \emph{Proceedings of the 2020 Conference on Empirical Methods in
  Natural Language Processing (EMNLP)}, pages 2685--2702, Online. Association
  for Computational Linguistics.

\bibitem[{Sennrich et~al.(2016)Sennrich, Haddow, and
  Birch}]{sennrich-etal-2016-neural}
Rico Sennrich, Barry Haddow, and Alexandra Birch. 2016.
\newblock \href {https://doi.org/10.18653/v1/P16-1162} {Neural machine
  translation of rare words with subword units}.
\newblock In \emph{Proceedings of the 54th Annual Meeting of the Association
  for Computational Linguistics (Volume 1: Long Papers)}, pages 1715--1725,
  Berlin, Germany. Association for Computational Linguistics.

\bibitem[{Tay et~al.(2021)Tay, Tran, Ruder, Gupta, Chung, Bahri, Qin,
  Baumgartner, Yu, and Metzler}]{tay2021charformer}
Yi~Tay, Vinh~Q Tran, Sebastian Ruder, Jai Gupta, Hyung~Won Chung, Dara Bahri,
  Zhen Qin, Simon Baumgartner, Cong Yu, and Donald Metzler. 2021.
\newblock Charformer: Fast character transformers via gradient-based subword
  tokenization.
\newblock \emph{arXiv preprint arXiv:2106.12672}.

\bibitem[{Vaswani et~al.(2017)Vaswani, Shazeer, Parmar, Uszkoreit, Jones,
  Gomez, Kaiser, and Polosukhin}]{vaswani2017attention}
Ashish Vaswani, Noam Shazeer, Niki Parmar, Jakob Uszkoreit, Llion Jones,
  Aidan~N Gomez, {\L}ukasz Kaiser, and Illia Polosukhin. 2017.
\newblock Attention is all you need.
\newblock In \emph{Advances in neural information processing systems}, pages
  5998--6008.

\bibitem[{Xue et~al.(2021)Xue, Barua, Constant, Al-Rfou, Narang, Kale, Roberts,
  and Raffel}]{xue2021byt5}
Linting Xue, Aditya Barua, Noah Constant, Rami Al-Rfou, Sharan Narang, Mihir
  Kale, Adam Roberts, and Colin Raffel. 2021.
\newblock Byt5: Towards a token-free future with pre-trained byte-to-byte
  models.
\newblock \emph{arXiv preprint arXiv:2105.13626}.

\end{thebibliography}
\bibliographystyle{acl_natbib}

\appendix

\section{Test for Information Leaks} \label{A:leak}
Our approach for testing for information leaks in a downsampling method trains for 5000 iterations in batches of 32. We use the Adam optimizer \cite{kingma2014adam}, with a learning rate of 1e-4. These numbers were determined empirically, based on the degree of separation seen from the accuracies of the leaking versus non-leaking tokens. The accuracies in Table \ref{tab:leaks} are obtained over 100 batches, or 3200 samples. 

\section{Main Experimental Details} \label{A:exp}
\subsection{Datasets}
Our experiments use the IWSLT2017 data\footnote{\url{https://sites.google.com/site/iwsltevaluation2017/TED-tasks}} for English--German and English--Arabic, using the test sets from 2010 and 2015 for validation and testing, respectively. For English--Turkish, we use the SETIMES2 dataset\footnote{\url{https://opus.nlpl.eu/SETIMES2.php}}, and WMT's \verb|newstest2017| and \verb|newstest2018| for validation and testing.\footnote{\url{http://data.statmt.org/wmt18/translation-task/dev.tgz}}
\subsection{Preprocessing}
We remove sentence pairs where the English sentence is longer than 256 characters from the training data. To keep our vocabulary size consistent across all languages, we tokenize according to UTF-8 bytes rather than using a character vocabulary.\footnote{Operating on the byte-level also follows \citet{tay2021charformer}, despite the name ``Charformer'' perhaps suggesting otherwise.} We also include subword-level models, with vocabularies generated with SentencePiece \cite{kudo2018sentencepiece}. We chose a vocabulary size of 16 thousand for English--German and English--Arabic, as that roughly corresponds to a downsampling factor of 4. For English--Turkish, we used a vocabulary size of 8 thousand.\footnote{We also tested a vocabulary size of 16 thousand, but this performed worse.}
\subsection{Hyperparameters}
We use the Transformer model with the same parameters as Transformer Base \cite{vaswani2017attention}. We train our models using AdamW \cite{loshchilov2017decoupled} with a learning rate of 2e-4, a linear warmup of 4000 steps, a batch size of 128, and label smoothing factor of 0.1. The learning rate was chosen from a grid search, the batch size chosen empirically, and the warmup steps and label smoothing factor were based on \citet{libovicky2021don}. We use an early stopping criterion of no improvement on the validation set with a patience of 10. All of our models are trained on a single Nvidia V100 (32GB) GPU.

\section{Comparison of GBST modifications on Arabic$\leftrightarrow$English.} \label{A:mods}
In addition to German$\leftrightarrow$English (Table \ref{tab:gbst}), we compare the GBST modifications on Arabic$\leftrightarrow$English, shown in Table \ref{tab:gbst_ar}. The results are more promising for the masking method compared to those on English$\leftrightarrow$German, however they are still outperformed by the removal method.
\begin{table}[!htp]\centering
\scriptsize
\begin{tabular}{lrrrrr}\toprule
$\delta$ & Non-Causal &Padding &Removal &Masking \\\midrule
2 & 0.00 & 0.80 &24.21 & 24.02\\
3 & 0.00 & 0.12 &23.70 & 22.50\\
4 & 0.00 & 0.02 &22.24 & 21.26\\ \midrule
2 & 0.00 & 0.04 &7.81 & 7.62\\
3 & 0.00 & 0.30 &7.77 & 7.21\\
4 & 0.00 & 0.23 &8.32 & 7.73\\
\bottomrule
\end{tabular}
\caption{Arabic$\rightarrow$English (top) and English$\rightarrow$Arabic (bottom). }\label{tab:gbst_ar}
\end{table}

\end{document}